\documentclass[10pt,twocolumn,letterpaper]{article}

\usepackage{cvpr}              

\usepackage{graphicx}
\usepackage{amsmath}
\usepackage{amssymb}
\usepackage{booktabs}
\usepackage{multirow}

\usepackage[accsupp]{axessibility}

\usepackage{hyperref}
\hypersetup{colorlinks,allcolors=black}

\usepackage[capitalize]{cleveref}
\crefname{section}{Sec.}{Secs.}
\Crefname{section}{Section}{Sections}
\Crefname{table}{Table}{Tables}
\crefname{table}{Tab.}{Tabs.}

\def\cvprPaperID{*****} 
\def\confName{CVPR}
\def\confYear{2022}

\begin{document}


\title{ActAR: Actor-Driven Pose Embeddings for Video Action Recognition}


\author{Soufiane Lamghari, Guillaume-Alexandre Bilodeau\\
LITIV lab., Dept. of Computer and Software Eng.\\
Polytechnique Montreal \\
Montreal, Canada\\
{\tt\small \{soufiane.lamghari, gabilodeau\}@polymtl.ca}
\and
Nicolas Saunier\\
Dept. of Civil, Geological and Mining Eng.\\
Polytechnique Montreal \\
Montreal, Canada\\
{\tt\small nicolas.saunier@polymtl.ca}
}
\maketitle

\begin{abstract}

Human action recognition (HAR) in videos is one of the core tasks of video understanding. Based on video sequences, the goal is to recognize actions performed by humans. While HAR has received much attention in the visible spectrum, action recognition in infrared videos is little studied. Accurate recognition of human actions in the infrared domain is a highly challenging task because of the redundant and indistinguishable texture features present in the sequence. Furthermore, in some cases, challenges arise from the irrelevant information induced by the presence of multiple active persons not contributing to the actual action of interest. Therefore, most existing methods consider a standard paradigm that does not take into account these challenges, which is in some part due to the ambiguous definition of the recognition task in some cases. In this paper, we propose a new method that simultaneously learns to recognize efficiently human actions in the infrared spectrum, while automatically identifying the key-actors performing the action without using any prior knowledge or explicit annotations. Our method is composed of three stages. In the first stage, optical flow-based key-actor identification is performed. Then for each key-actor, we estimate key-poses that will guide the frame selection process. A scale-invariant encoding process along with embedded pose filtering are performed in order to enhance the quality of action representations. Experimental results on InfAR dataset show that our proposed model achieves promising recognition performance and learns useful action representations.

\end{abstract}

\section{Introduction}

Human Action Recognition (HAR) is a fundamental research problem in computer vision that aims to categorize human actions. This task has seen a lot of advances in recent years, making it relevant to a wide range of applications, such as surveillance and security, human robot interaction, autonomous vehicles, and urban planning ~\cite{sudha2017approaches, xia2015robot, lu2020driver}. Human action recognition is a very complex task due to several challenges such as intra-class variations, viewpoint variations, motion velocity variations, background clutter, and human body occlusions.

\begin{figure}[t]
    \centering
    \includegraphics[scale=0.4]{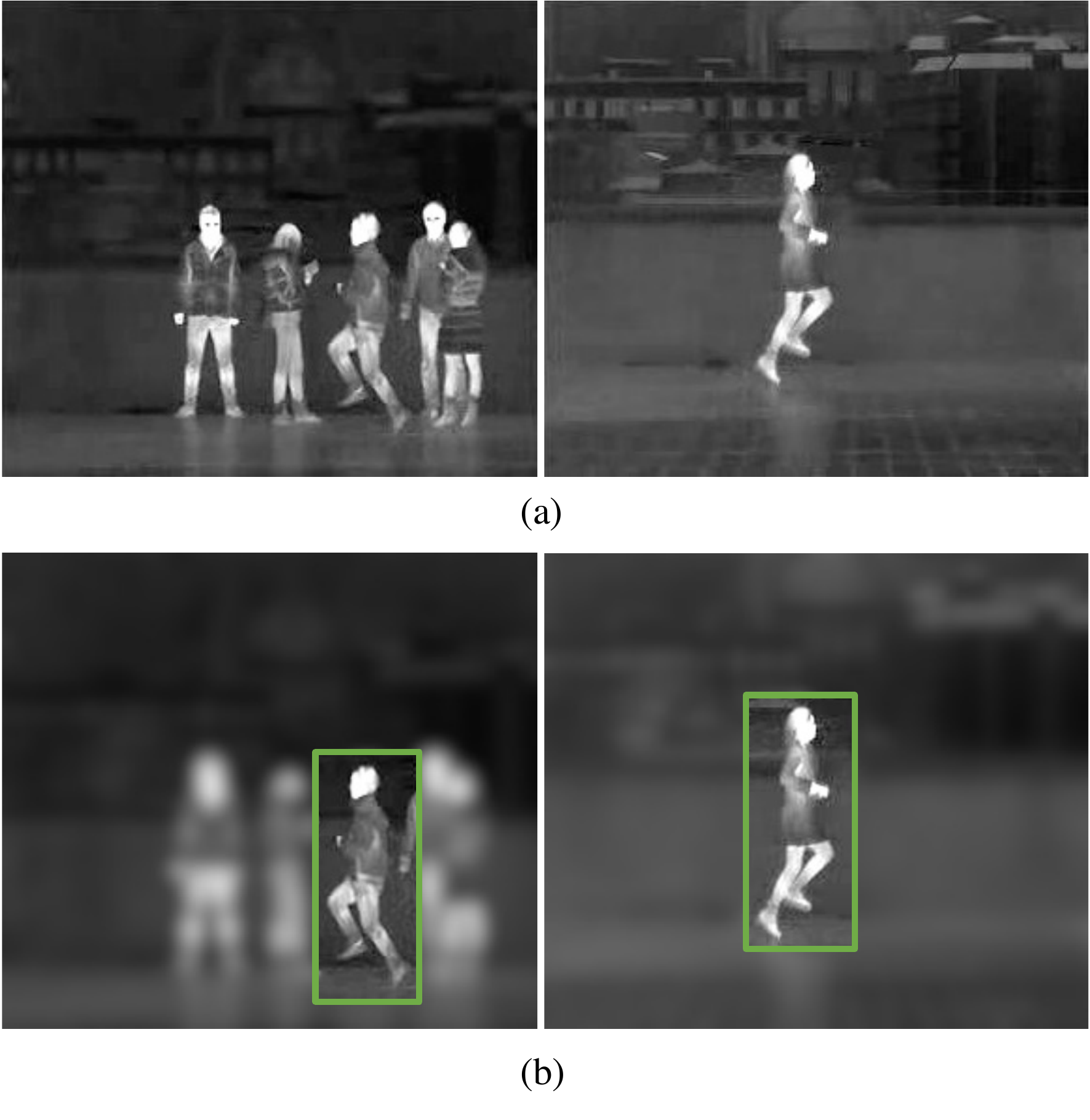}
    \caption{Example of two actions involving single and multiple persons from the InfAR dataset labeled as "Skip". (a) Considering all people in a scene to recognize an action can be uninformative. (b) Identifying the key-actor performing the main action provides accurate guidance for the action recognition model.}
    \label{fig:1}
\end{figure}

Thanks to the advances in deep learning, several methods exploiting convolutional neural networks (CNNs) for human action recognition in the visible domain have been proposed~\cite{simonyan2014two, tran2015learning, donahue2015long, feichtenhofer2016convolutional, wang2016temporal, carreira2017quo}. Compared to classical approaches, such as dense trajectories~\cite{wangh_dense_traj2011}, the two-stream CNN architecture~\cite{simonyan2014two} has improved action recognition performance by fusing the output of separate spatial and temporal networks. The first stream network learns spatial salient appearance patterns and the second stream network learns temporal motion features. Wang et al.~\cite{wang2016temporal} demonstrated that dividing each video sequence into multiple segments, each processed with a two-stream network, and aggregating their classification scores yields even better recognition results. To enhance human action recognition performances, several other works~\cite{cheron2015p, zolfaghari2017chained, lamghari2021grid, yan2018spatial, shi2019two, asghari2020dynamic} have investigated the integration of human pose sequences or skeletal data information as they provide robust and discriminative representations of actions. 

Despite the success of current state-of-the-art methods on visible videos, they still show limited performance when applied in certain real-world situations with low-light conditions such as fog, smoke, or even dark environments. In such scenes, infrared videos are more suitable as they are resistant to background clutter, and less sensitive to lighting conditions. Few works have addressed action recognition in infrared videos~\cite{zhu2013study, gao2016infar, jiang2017learning}. Gao et al.~\cite{gao2016infar} proposed a two-stream CNN framework for infrared action recognition. Jiang et al.~\cite{jiang2017learning} passed infrared data and optical flow volumes to a two-stream 3D CNN framework and integrated a discriminative code layer to generate class-based representations. 

While these methods have mainly focused on incorporating richer information (e.g., original images, optical flow, motion-history images of optical flow (OF-MHI)~\cite{bobick2001recognition}) by increasing the number of input streams, they are not robust enough to provide discriminative action representations. This is mainly due to three reasons. First, they do not differentiate multi-person videos
from single-person videos when recognizing actions, and learn global scene-level features without focusing on key-actors performing the main action. In fact, videos captured in realistic environments usually contain multiple persons interacting with each other, but only a subset of them at a given time is involved in the main action, as exemplified by the infrared action recognition dataset, InfAR~\cite{gao2016infar}. For instance, a ``Skip'' action in a scene is defined by one human (see Figure~\ref{fig:1}). Thus, integrating spatial and temporal information about other people not involved in the main action, can be uninformative and misleading and prevent the model from learning specific cues to understand such an action. Second, they learn redundant or random features instead of focusing only on relevant frames. Third, these models often consider the stream networks to be independent and do not allow information sharing between them. They encode only the motion features in short time windows without guaranteeing the preservation of the discriminative cues with pooling techniques.

To address the above shortcomings, we propose Actor-Driven Pose Embeddings for Video Action Recognition (ActAR), a novel model for actor-specific action modeling and recognition in the infrared spectrum. Our model operates in a three-stage fashion. First, we identify key-persons in a video performing an action based on optical flow, and we ignore the others that do not contribute to the main action. This allows our model to focus only on relevant scene information that characterize informative human actions. In the second stage, we extract body pose information for each key-actor. By considering pose information, we provide rich cues to compensate for the missing texture and color information in the infrared spectrum. As most pose estimation models are trained in the visible spectrum, they can fail in some cases when applied directly to the infrared domain. To solve this problem, we propose a pose embedding filtering mechanism that can leverage complementary discriminative capabilities from large-scale RGB datasets. From the filtered pose sequences, we extract key-poses to restrict the analysis to only the relevant temporal action features. Last, we integrate the most representative features into a compact representation that encodes both the infrared and pose information to obtain a more relevant representation of the performed action. We tested our ActAR approach on the InfAR dataset and found that it achieves good improvement on human action recognition tasks.

Our contributions can be summarized as follows:

\begin{itemize}
    \item We introduce a unified model for scene understanding by simultaneously addressing two tasks in a single framework: main actor identification and human action recognition.
    
    \item Our method operates on raw infrared video sequences and identifies key body pose features for robust and discriminative action representations. 

    \item Extensive experiments on the challenging InfAR dataset demonstrate the effectiveness of our proposed model compared to the state-of-the-art.

\end{itemize}

\section{Related work}
In this section, we first review existing works on human action recognition in videos, then discuss the importance of pose features for human action recognition.

\subsection{Human Action Recognition in Videos}

Human action recognition in videos aims to classify videos into pre-defined sets of actions. Research methodologies in this field can be subdivided into two main categories: handcrafted features-based and deep learning-based approaches. Early models extracted handcrafted features with a series of operations, such as translations, rotations and computing the trajectories of points to solve HAR related problems. In \cite{weinland2006free}, a viewpoint-free representation for human action from view-invariant motion descriptors was proposed, where only variations in viewpoints related to the central vertical axis of the human body are considered. Later, Wang and Schmid~\cite{wang2013action} proposed to estimate camera motion to improve the dense trajectories video representation for human action recognition. Cov3DJ~\cite{hussein2013human} uses a covariance matrix of skeletal joint movements as descriptors combined with a classification algorithm to recognize human actions. Despite their success, handcrafted methods capture only local features, making them less discriminative to correctly recognize complex human actions.

Deep-learning based approaches are now dominating this field, as their performance is significantly better compared to handcrafted-based approaches for HAR. In this context, multiple research works have proposed to exploit CNN models for action recognition~\cite{simonyan2014two, tran2015learning,carreira2017quo}. The work of~\cite{simonyan2014two} is notable in this category as it has successfully integrated CNNs for human action recognition by introducing a two-stream method. This method relies on RGB and optical flow and models them into a spatial stream that carries the scene and target information in the video, and a temporal stream that focuses on the target and camera motion. Tran et al.~\cite{tran2015learning} proposed C3D, a method that simultaneously models appearance and motion information using 3D ConvNets and successfully demonstrated that 3D ConvNets are more appropriate for spatio-temporal feature learning than 2D ConvNets. Two-Stream Inflated 3D ConvNet (I3D) was later proposed in~\cite{carreira2017quo}. It inflates filters and pooling kernels used for deep image classification ConvNets into 3D, leading to natural spatio-temporal feature extraction from videos.

Since RNNs are suitable for sequential data modeling, various RNN-based methods for human action recognition were proposed. For instance, Srivastava et al.~\cite{srivastava2015unsupervised} train LSTMs with self-prediction to learn salient video representations, while Gammulle et al.~\cite{gammulle2017two} created a deep fusion framework by learning spatial features from different layers of CNNs and then mapping them with temporal features from LSTMs. Later, Du et al.~\cite{du2017recurrent} integrated an attention mechanism into a recurrent spatial-temporal network to learn context key features for each timestep prediction of RNN. However, RNN-based methods usually suffer from  gradient vanishing and nonparallelism issues making them hard to optimize~\cite{pascanu2013difficulty}. 

Compared with HAR in the visible domain, very few works have addressed this task in the infrared spectrum regardless of its great potential in handling low-light conditions~\cite{zhu2013study, gao2016infar, jiang2017learning}. Gao et al.~\cite{gao2016infar} proposed a two-stream CNN framework for infrared action recognition. Jiang et al.~\cite{jiang2017learning} passed infrared data and optical flow volumes to a two-stream 3D CNN framework and integrated a discriminative code layer to generate class-based representations. Later, several multi-stream methods were also proposed. For instance, Liu et al.~\cite{liu2018global} adopted a three-stream framework over local, global, and spatio-temporal features to learn discriminative action representations. While these methods have mainly focused on incorporating extra information (e.g., original images, optical flow, motion-history images of optical flow (OF-MHI)~\cite{bobick2001recognition}) by increasing the number of input streams, they are still not robust enough to provide discriminative action representations.

\subsection{Pose-based Action Recognition}

Pose-based human action recognition has recently attracted considerable interest in the literature, as the human body joints position and motion are shown to provide discriminative cues for action recognition~\cite{jhuang2013towards}. Cheron et al.~\cite{cheron2015p} proposed an action descriptor conditioned on motion and appearance CNN features computed for all body parts,  while Zolfaghari et al.~\cite{zolfaghari2017chained} combine pose, motion and RGB information using a Markov chain model to classify actions. Similarly, a multitask framework with a single architecture was proposed in~\cite{luvizon20182d} for pose estimation and human action recognition. Choutas et al.~\cite{choutas2018potion} suggested to encode long-term dependencies of pose keypoints motion and combine it with a shallow CNN to classify actions. Asghari et al.~\cite{asghari2020dynamic} later improved upon this
by inferring a fixed-length representation based on the location and body joints heatmap. Unlike previous methodologies, this paper presents a different architecture for HAR that extracts richer key-poses representation.

Besides body pose information, skeleton data consisting in joints and bones representation of the human body, was also considered in the context of human action recognition~\cite{yan2018spatial, shi2019two, wei20203d}. Skeletal body joints information can be acquired using specific sensors such as depth cameras. However, the associated cost and availability of such data makes it limited in practice. Thus, such solutions are outside the scope of this paper.

\section{Proposed method}

In this section, we introduce our proposed Actor-Driven Pose Embeddings for Video Action
Recognition (ActAR) method. Specifically, we first present our new key-actor identification module, which defines the main actor performing the action. We then
introduce the building blocks of ActAR, including human pose estimation, embedded pose filtering, key-pose selection, and human action learning.

\subsection{Overview}

\begin{figure*}
    \centering
    \includegraphics[scale=0.35]{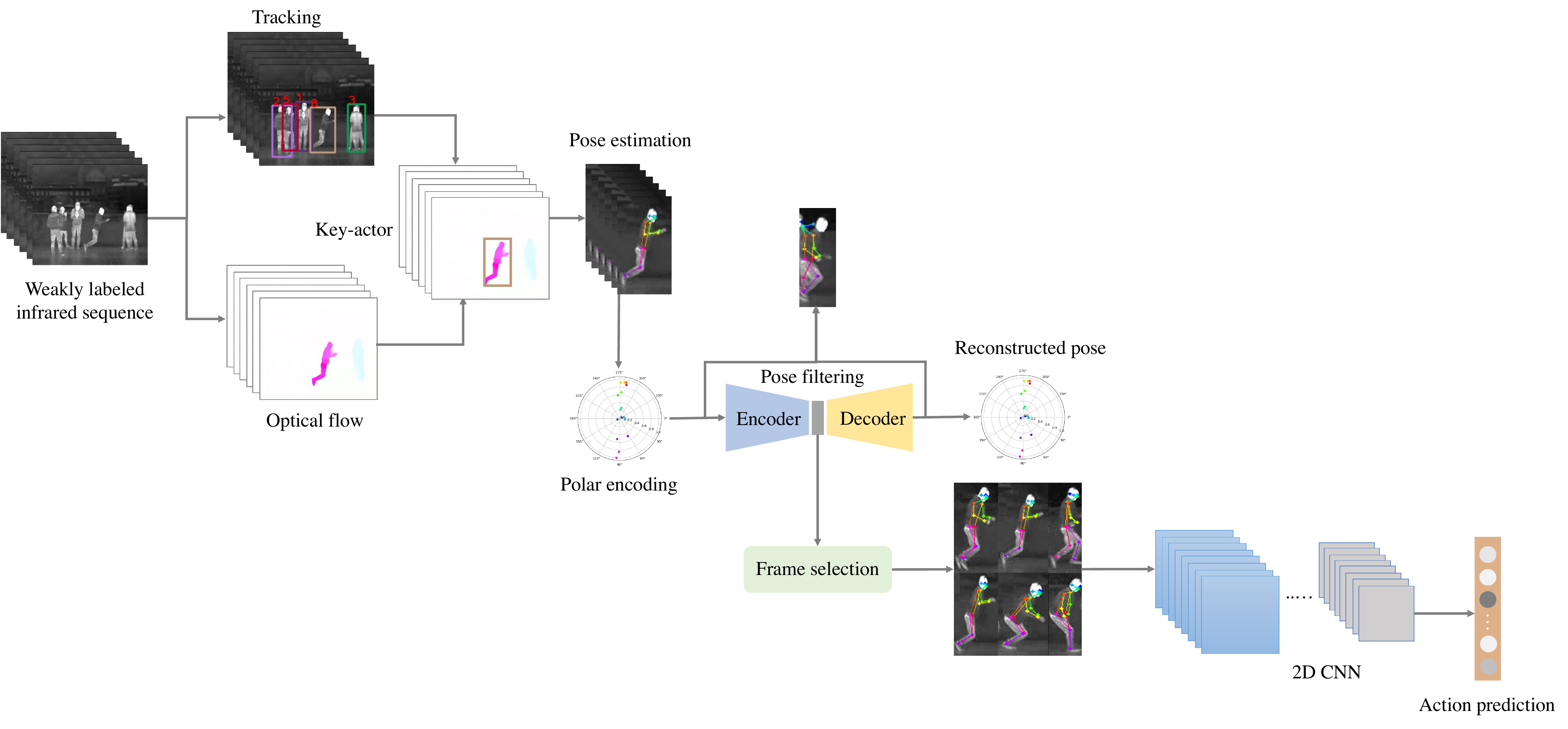}
    \caption{An overview of the proposed ActAR model. Given an infrared input video, we identify the key-actors based on optical flow. To compensate for the missing texture and color information in the infrared domain, we extract the human body keypoints. We project the estimated keypoints into the polar system, and then we filter out incorrect poses that may mislead the action representation. Next, we select the key-poses based on a deep clustering method to form a compact representation encoding the most relevant infrared and pose information. In the last stage, we train a 2D deep convolutional neural network to predict the human action.}
    \label{pipeline}
\end{figure*}

The proposed approach for human action recognition in videos consists of four components, as illustrated in Figure~\ref{pipeline}. First, the main actors in each sequence are identified using motion information. This helps us focus on regions of interest in each frame, rather than analyzing irrelevant features present in each frame. Second, we extract pose features for each identified actor, as human poses are regarded as a very discriminative and distinctive information to recognize actions. Incorrect pose estimations are filtered through a deep pose embedded filtering mechanism to avoid introducing inaccurate information in the learning process. Based on the filtered pose features, we select the most discriminative key-poses that are sufficient to describe the main posture states of the actor while performing the action. By representing each actor with a set of key-poses, we optimize the learning process by providing only the relevant key features and disregarding the redundant ones. Third, to make the extracted key-poses amenable to the learning space, we embed them in a compact representation that summarizes the distinctive key-poses of each actor along with the infrared information that provides extra context to the keypoint features. Finally, we train a CNN with the constructed compact representations to learn and recognize human actions in video sequences.

\subsection{Key-Actor Identification}

Human action recognition in multi-person videos requires a deep understanding of the scene features. However, in weakly-labeled real-world datasets, scenes can contain multiple persons interacting with each other, but not necessarily all of them are involved in the main action. This makes the task of human action recognition in such situations even more challenging. One solution to this problem would be to focus only on the persons performing the labeled action. Such annotations are expensive to acquire and are not always readily available. Motivated by the need for automated key-actors identification in realistic scenes, we propose a new methodology for actor-specific action recognition that identifies and locates the main person performing the action.

The key-actor identification task requires accurate human body detection and tracking despite all the challenges imposed by realistic videos in the infrared spectrum. To track humans in videos, we apply a multi-object tracker over each frame $f$ of the video sequence $S$ of $T$ frames. We adopted the recent state-of-the-art Bytetrack model~\cite{zhang2021bytetrack} to track simultaneously multiple human instances in videos. Given the extracted tracks $[Tr_a,...,Tr_A]$, where $Tr_{a}=[b_{a0},...,b_{aT}]$. The challenge now is to identify the main actor tracks as the other tracks are not informative to the action and may introduce noisy and non-discriminative features for the action recognition model. One robust information that can assist the identification of the main actor is motion feature. At each timestep, and for each track, we extract the optical flow magnitude in the object bounding box $y(Tr_a)=[y_{a0},...,y_{aT}]$. To achieve this, we use the GMA model~\cite{jiang2021learning}. This method is able to handle occlusions by aggregating the global motion using a transformer. The aggregated motion features are defined as:

\begin{equation}
    \hat{y_i} = y_i + \alpha \sum_{j=1}^{N} f( \theta(x_i), \phi(x_j)) \sigma(y_i))
\end{equation}

\noindent where $x$ denotes the context features, $y$ are the motion features, $f$ refers to a similarity attention function, $\theta$, $\phi$ and $\sigma$ are the projection functions for the query, key, and value vectors, respectively, and $\alpha$ is a scalar parameter.

Note that frames with low-score tracking are discarded. To distinguish the main actor, we maximize the motion vector of all actors, and choose the one with strongest motion information, where the key-actor id is defined by: 

\begin{equation}
    k_{id}= \text{argmax} \big(y(Tr_a),...,y(Tr_A)\big)
\end{equation}

Now that we have performed the first precision-level of our model by identifying the temporal tracks of the main actor, we next discuss how to estimate and select discriminative infrared information in the temporal dimension. 

\subsection{Human Pose Estimation}

In this work, we mainly guide the task of action recognition from videos by using the evolution of human body poses as a cue. These representations will enable us to deal with the abrupt viewpoint variation and lighting change challenges. These representations are also compact and low-dimensional, which is an essential feature for optimized video analysis. However, for several benchmark datasets, human pose information is not readily available. To address this problem, we estimate 2D human pose features of the identified actors in each video frame using the state-of-the-art human pose detector proposed in~\cite{sun2019deep}. The High-Resolution Net (HRNet) model learns reliable pose estimation of multiple persons in videos using a high-resolution network. It starts with a high-resolution subnetwork, then high-to-low resolution subnetworks are added subsequently, while multiscale fusions are performed repeatedly to allow information sharing between all the parallel multi-resolution subnetworks.

Formally, for a given video sequence $S \in \mathbb{R}^{T \times  H \times W \times 3}$ of $T$ frames with width $W$, and height $H$, we run the HRNET pose detector on each frame to get human pose keypoints (i.e., body joints). Thus, across all frames, each actor $h$ is represented by a set of $K$ temporal sequence of keypoints denoting the location estimate of each body joint of the main actor at every frame. During the pose estimation process, we optimize the MSE loss function as follows:

\begin{equation}
    \mathcal{L}_{mse} = \frac{1}{M} \sum_{m=1}^{M} \| C_m - \hat{C}_m  \|^2_2 
\end{equation}

\noindent where $C_m$ and $\hat{C}_m$ refer to the true and the predicted confidence map for the $m^{th}$ body joint, respectively.

This module returns the evolution of each actor body pose across time, which will be used by our model for human action recognition. Next, we encode the extracted pose features to mitigate the effects related to the camera viewpoint changes that may lead to observing the same person with different appearance scales.

\subsection{Scale-invariant Polar Encoding and Embedded Pose Filtering}

The estimated human body poses often contain large variations, such as scale, and bone length ratios~\cite{liu2021normalized}. This leads to having the same action represented by different spatial keypoints coordinates and thus can mislead the learning model. To alleviate this, we propose a scale-invariant encoding method to eliminate the influence of such effects. The goal is to let the pose-guided features encode only the human poses related to the action while being invariant to other factors. Specifically, we select the center hip joint as the reference body joint, because it is usually stable in most actions. Then, we calculate the relative positions of all the other joints with respect to the reference joint $\hat{J^{t}}$:

\begin{equation}
    J_{m}^{t} = J_{m}^{t} - \hat{J^{t}}
\end{equation}

As the actor body joints are initially defined in the Cartesian coordinates system $(x,y) \in \mathbb{R}^2$, we next project them to the polar coordinate system with a polar angle $\theta$ $\in$ $[0, 2\pi]$. This enables our model to differentiate between keypoint vectors using their angles instead of focusing on their magnitude. Thus, in this new system, the hip center becomes the pole.

In the same context of enhancing the quality of the estimated poses, we have noticed that when using a model trained on RGB data, the predictions on infrared domain may result in incorrect poses in some cases. To overcome this problem, we propose an embedded pose filtering mechanism that leverages latent pose representations from the visible domain to filter poses estimated in the infrared spectrum. To do so, we train a deep autoencoder model to reconstruct ground truth poses from COCO dataset~\cite{lin2014microsoft}. The trained autoencoder will be also used in the clustering module as we will discuss in the next section. Once the autoencoder has learned the latent pose representations in the rgb domain, the model decides to keep or discard the estimated infrared pose based on the reconstruction error.

\subsection{Human Pose Features Selection}
\label{sect:clustering}

Since human poses in consecutive frames are often extremely similar, we select individual distinct key-poses that are used as representative poses summarizing the principal actor actions. Focusing on frames with key-poses significantly reduces the complexity and the computational cost of the human action recognition algorithm. To infer these candidate key-poses, we embed the extracted pose features of every actor across frames into a latent space and cluster the resulting embedded vectors into sets of similar poses. Thus, we adopt a Deep Embedded Clustering method~\cite{xie2016unsupervised} as a base model for our key-pose extraction process. This model consists in a deep autoencoder, followed by a soft clustering layer that assigns each pose sample to a cluster with a specific probability. Specifically, it uses the autoencoder networks to embed input features into a low-dimensional latent space, and infers cluster assignments simultaneously.

Formally, given a polar encoded-keypoints sequence $V_h$ for a target actor, we start by learning unsupervised representations of the input keypoints using a stacked autoencoder (SAE). It has been shown that this type of networks learns discriminative and meaningful latent representations on various datasets~\cite{le2013building, xie2016unsupervised}. After initializing the network,  we remove the decoder to keep just the encoder for deep features extraction. After getting the embedded pose keypoints, we perform standard k-means clustering to obtain the initial cluster centers. And then we optimize the encoder using the following clustering objective:

\begin{equation}
    L = KL(P || Q )  = \sum_{i}  \sum_{j} p_{ij} \log \frac{p_{ij}}{q_{ij}}
\end{equation}

This objective function calculates the Kullback–Leibler (KL) divergence loss between the auxiliary target distribution $p_{ij}$ and the soft assignments $q_{ij}$ where,

\begin{equation}
    p_{ij} = \frac{ q^{2}_{ij} / \sum_{i} q_{ij}}{\sum_{j} q^{2}_{ij} / \sum_{i} q_{ij}}
\end{equation}

\noindent and $q_{ij}$ refers to the similarity between embedded pose keypoints $V'_h$ and cluster centroid $\mu_j$ measured by the Student's t-distribution~\cite{van2009learning}.

\begin{equation}
    q_{ij} = \frac{(1 + ||V'_{h} - \mu_j||^2)^{-1} }{\sum_{j} (1 + ||V'_{h} - \mu_j||^2)^{-1}}
\end{equation}

It is to note that minimizing the objective $L$ follows a self-training strategy, as the auxiliary target distribution $p_{ij}$ depends on the soft assignments $q_{ij}$ leading to enhanced clustering predictions.

Finally, to select the actual representative poses of each cluster, we calculate the distance between each pose $p_i$ and the embedded cluster center $p_{ic}$, and we take the nearest-neighbor pose sample to the cluster center as the representative key-pose for that cluster. We empirically found that the optimal number of clusters is 8, as it provided the best balance between too few and too many poses clusters. Thus, each actor in a sequence is mainly represented by 8 identified key-poses.

\subsection{Human Action Representation Learning}

To enhance the action recognition performance, various fusion techniques to aggregate temporal information with other modalities have been studied in the literature~\cite{karpathy2014large, simonyan2014two}. However, most of these approaches rely on information that is not always relevant for the human recognition task, and are limited in the way they fuse the temporal information. Unlike these methods, our ActAR model combines actor-specific relevant infrared information with the estimated key-poses into a compact representation. This enables our model to learn key patterns needed to recognize human actions in infrared videos.

To construct the summarized action representation for each key-actor, we follow these steps: we first extract the infrared region of interest $I_{hk}$ that is defined by the bounding box of each key pose in $V_{h}^{*}$. By encoding the pose vector directly for the selected relevant actor region, we compensate for the missing texture and color information in infrared videos.

Next, we aggregate each encoded region $I_{hk}$ into one main grid-like structure for every identified key-actor as illustrated in Figure~\ref{pipeline}. To prevent the model from learning unnecessary action patterns, we combined each encoded region with zeros-valued borders having a 3 pixel width, instead of being completely adjacent, as the the convolution kernel sizes of the adopted CNN architecture can be larger than the inter-region spacing.

After constructing the compact features representation for each key-actor, we train a convolutional neural network to learn the key patterns related to the performed action from the created grid structures. As our backbone CNN architecture, we use the Inception-ResNet-v2 model~\cite{szegedy2017inception}, which has been pre-trained on the ImageNet dataset~\cite{deng2009imagenet}. The Inception-ResNet-v2 model has a hybrid Inception network architecture that uses residual connections instead of filter concatenation. For the training objective, we adopt the categorical cross entropy loss defined as: 

\begin{equation}
    \mathcal{L}_{CE} = - \frac{1}{N} \sum_{i=1}^{N} \log \frac{\exp{({w_{y_i}^{T}G_i + b_{y_i}})}}{\sum_{j=1}^{n} \exp{({w_{j}^{T}G_i + b_{j}})}}
\end{equation}

\noindent where $G_i$ denotes the $i^{th}$ grid-like structure, $N$ is the number of training structures, $y_i$ refers to the class label of $G_i$, $W$ is the learned weight matrix, and $b$ is the intercept.


\begin{table*}
  \resizebox{\textwidth}{!}{
  \begin{tabular}{c c c c c c | c}
    \hline
      \multicolumn{2}{c}{Actor Identification} &
      \multicolumn{2}{c}{Frame Selection} &
      Polar Encoding &
        Pose Embedding Filtering &
            Average Precision (\%) \\
      \cline{1-2}     \cline{3-4}
  All actors & Principal actors & Random & Key frames &  &  &  \\
    \hline \hline 
    \checkmark &  & \checkmark  &  &  &  & 74.44 \\
     & \checkmark & \checkmark &  &  &  & 79.38 \\
     & \checkmark &  & \checkmark &  &  & 82.55 \\
     & \checkmark &  & \checkmark & \checkmark &  & 84.26 \\
     & \checkmark &  & \checkmark & \checkmark & \checkmark & 85.32 \\
    \hline
  \end{tabular}
  }
  \caption{Performance contribution of each component in ActAR on InfAR dataset.}
  \label{tab:abl1}
\end{table*}

\section{Experiments}

After describing the experimental settings, we first present an ablation study and then a comparison with the state-of-the-art.

\subsection{Experimental Settings}

All experiments were performed on a single TITAN Xp GPU. We implemented our proposed model using the TensorFlow library~\cite{abadi2016tensorflow}. We use Inception-ResNet-v2~\cite{szegedy2017inception} as our CNN backbone pre-trained on the ImageNet dataset~\cite{deng2009imagenet}. We trained the model using a Stochastic Gradient Descent (SGD) with ADAM~\cite{kingma2014adam} and set the optimizer hyperparameters to $\beta_1$ = 0.9, $\beta_2$ = 0.999, $\epsilon$ = 0.001. For the key-poses, we used  8 components and trained the network with a learning rate of $10^{-3}$ for 150 epochs with a mini batch size of 32.

For optical flow estimation we used the GMA method~\cite{jiang2021learning} trained on Sintel dataset~\cite{butler2012naturalistic}. For human tracking, we use the ByteTrack model proposed by Zhang et al.~\cite{zhang2021bytetrack}. We use the HighResolution Net (HRNet) architecture~\cite{wang2020deep} to compute human poses across frames. Specifically, we used the pose-hrnet-w48 architecture trained on the COCO dataset~\cite{lin2014microsoft}.

We evaluated the proposed Actor-Driven Pose Embeddings model on the Infrared human action recognition dataset, InfAR~\cite{gao2016infar}. Infrared Action Recognition (InfAR) is a challenging video action recognition dataset that provides 600 video clips captured by infrared thermal imaging cameras. Each video clip is annotated with one action from 12 different human action classes. The average video clip length is 4 seconds with a frame rate of 25 and a resolution of 293$\times$256.

\subsection{Ablation Study}

In this section, we perform an extensive ablation study to investigate the effectiveness of each component of our method on the InfAR dataset. In all ablations, we use average precision as the main evaluation metric. We report the ablation performance of our model in Table~\ref{tab:abl1}.

We start with a basic model to understand the impact of the actor identification stage. In this experiment, we first consider all persons in the scene as being involved in the main action, and we select random frames for action representation for each person. To infer the scene action classification, we take the most frequent of all person-level actions. With the basic model, we achieve 74.44\%. This is expected as the model considers irrelevant information at two levels of learning, namely, the scene and the actor levels.

To avoid incorporating wrong action information induced by persons who do not contribute to the main action in the scene, we integrate in the basic model information solely about key-actors using our proposed optical flow-guided module. Similar to our previous experiment, we randomly select frames to represent the key-actor action. With this model configuration, we outperformed the basic model with a precision of 79.38\%. This demonstrates that the same model can get better performance if it is fed with the right features with respect to the key-actor involved in the labeled action.

Next, we upgrade the key-actor model with key-poses-based frames selection. As we previously explained, key-poses are obtained by clustering key-actor body poses using a deep autoencoder model. Clustering performance will be discussed later in this section. Compared with random frames representations, choosing infrared information guided by key-poses to encode action cues provides an improved level of precision in the temporal domain and allows our model to capture more discriminative action features. The obtained results clearly highlight the advantage of incorporating the frame selection module.

We also studied the impact of 2D-keypoint representations on the final action recognition precision. While in the previous experiments, we used actor body joints defined in the Cartesian coordinates system, they are now projected to the polar coordinate system with the center hip as a reference joint since this joint is stable in most actions. Having human poses $\pi$-scaled in this new coordinate system significantly benefits the clustering module and enhances the recognition capabilities of our model, as shown in the results. The new keypoints are not only lying on a common space but are also scale-invariant.

As the InfAR dataset is not manually pose-labeled, the keypoints predicted by HRNet can be sometimes incorrect. The integration of all the previous model components with the pose embedding filtering module, i.e., our ActAR model, mitigates the effects of this problem, and thus, reaches the highest action recognition performances. The performance boosts induced by the pose embedding filtering module indicate the advantage of filtering features predicted by trained RGB models in the infrared domain, especially when working with weakly labeled data.

\begin{figure}
    \centering
    \includegraphics[scale=0.55]{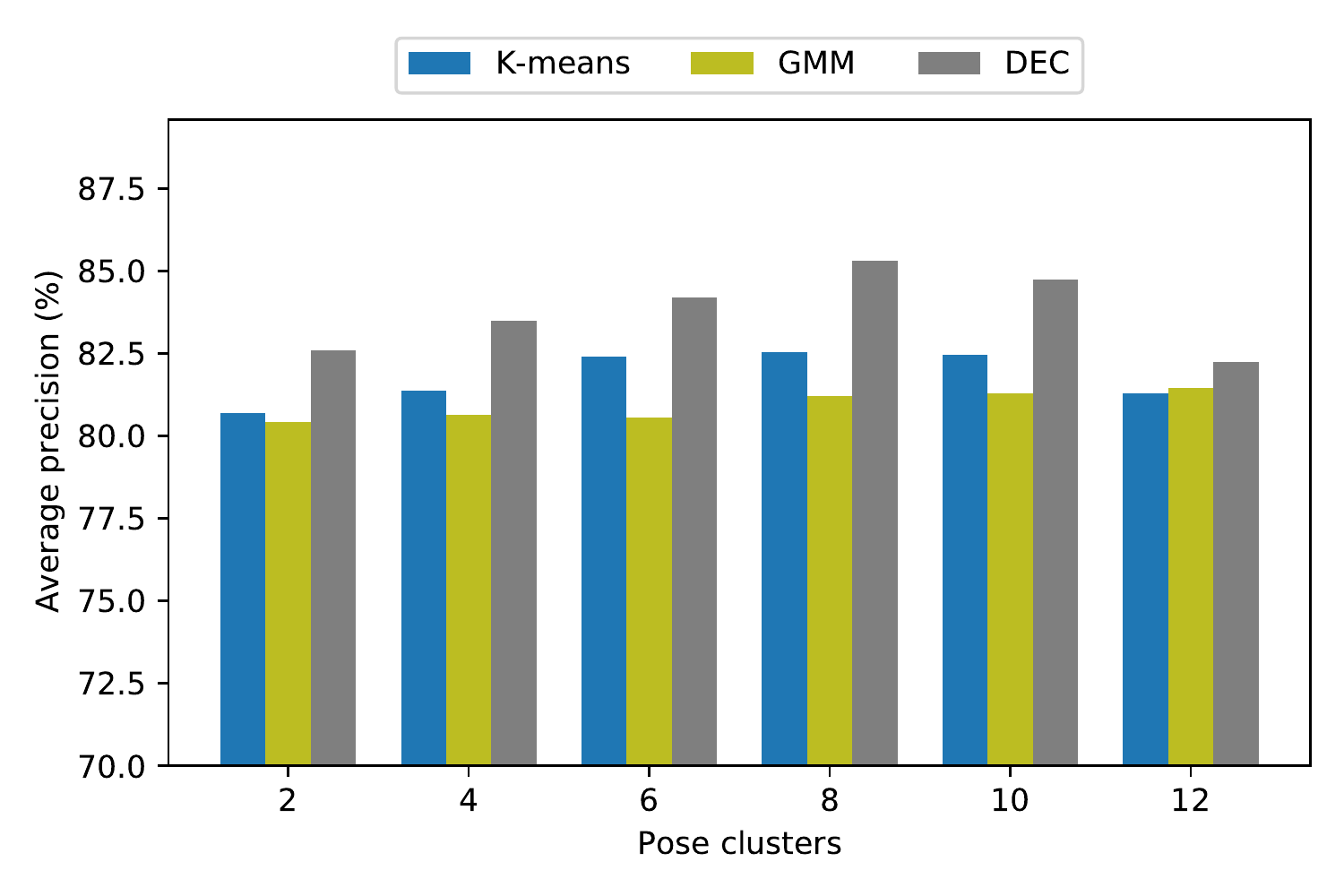}
    \caption{Impact of different pose clustering algorithms on the performance of ActAR on InfAR dataset.}
    \label{fig:clust_ablation}
\end{figure}

We have seen above that frame selection guided by key-poses clustering can improve the performance of our model for human action recognition. Now, we want to evaluate the impact of the clustering model for identifying the most informative body poses for each action. We compared three clustering methods: K-means, Gaussian Mixture Model (GMM), and Deep Embedded Clustering (DEC) with different numbers of clusters, varying from 2 to 12. In this experiment, we use the complete version of ActAR with the same parameters settings. We report the clustering ablation results in Figure~\ref{fig:clust_ablation}. We can see that integrating DEC with ActAR achieves the highest precision. This demonstrates that clustering deep pose features is better than clustering raw keypoints coordinates. On the other hand, ActAR with GMM performs slightly worse than the k-means version in this experiment. This can be explained by the fact that the nature of our infrared input pose data is not naturally normally distributed. Furthermore, we found that the optimal number of key-poses is 8. This indicates that the number of key-poses must be chosen based on the average size of humans in the scene.

\subsection{Comparison with the State-of-the-Art Methods}

We compare the proposed ActAR model with the state-of-the-art methods on the InfAR dataset, which can be grouped into two categories: handcrafted features-based methods (e.g., DT, and iDT) and deep learning-based approaches (e.g., HOF, Two-stream 2D-CNN, Two-stream 3D-CNN, TSTDDs, Four-stream CNN, and SCA). 

The results of our model and existing methods are reported in Table~\ref{comp_sota_infar} using the average precision evaluation metric for human action predictions. We observe that our method achieves the best action recognition precision. In fact, ActAR surpasses handcrafted features-based methods~\cite{wangh_dense_traj2011,wang2013action} that uses dense trajectories with a considerable margin of 13.49\% and deep learning-based methods ~\cite{gao2016infar,jiang2017learning,liu2018global,imran2019deep,chen2021infrared} that uses different multi-stream architectures with a margin ranging from 1.07\% to 8.66\%. Our results support our hypothesis that our main actor-driven ActAR model with discriminative pose-based features is beneficial for the human action recognition task. The results also reveal the impact of the embedded pose filtering module on the recognition performance, especially when extracting features from infrared videos through models that are trained on labeled RGB data.

\begin{table}
    \centering
    \begin{tabular}{l | c}
    \hline
    Model    &  Average precision(\%) \\
    \hline \hline
    HOF~\cite{gao2016infar}    &  68.58 \\
    DT~\cite{wangh_dense_traj2011}      &  68.66 \\
    iDT~\cite{wang2013action}      &  71.83 \\
    Two-stream 2D-CNN~\cite{gao2016infar}     &  76.66 \\
    Two-stream 3D-CNN~\cite{jiang2017learning}     &  77.50 \\
    CDFAG~\cite{liu2018transferable}     &  78.55 \\
    TSTDDs~\cite{liu2018global}     &  79.25 \\
    Four-stream CNN~\cite{imran2019deep}     &  83.50 \\
    SCA~\cite{chen2021infrared}     &  84.25 \\
    \hline
    ActAR &  \textbf{85.32} \\
    \hline

 \end{tabular}
    \caption{Comparison of the average recognition precision  of ActAR with state-of-the-art methods on InfAR dataset.}
    \label{comp_sota_infar}
\end{table}

\section{Conclusion}

In this paper, we have introduced a new actor-driven model for effective human action recognition in infrared videos with single and multiple persons. Extensive evaluation experiments demonstrate the effectiveness of ActAR compared to existing methods, and highlights the impact of each component of the proposed model. The proposed actor identification and frame selection modules effectively encode discriminative infrared information with key-actors body poses, and combining them together in a compact representation leads to considerable improvements. This further shows that identifying key-actors responsible for the action is crucial for efficient human action recognition. Therefore, our model can be easily generalized to any multi-person sequence, as it does not require key-person annotations for training. Finally, The embedded pose filtering module also shows that infrared spectrum can benefit from RGB models and data by getting better quality features. 


\section*{Acknowledgment}
This work was supported by the National Sciences and Engineering Research Council of Canada (NSERC). We acknowledge the generous support of a Titan X GPU from NVIDIA Corporation.

{\small
\bibliographystyle{ieee_fullname}
\bibliography{egbib}
}

\end{document}